\documentclass[10pt,twocolumn,letterpaper]{article}

\usepackage[pagenumbers]{cvpr} 

\usepackage{makecell}

\usepackage{multirow}
\usepackage{colortbl}
\usepackage{float}
\definecolor{LightGray}{gray}{0.95}
\definecolor{SemiLightGray}{gray}{0.8}

\definecolor{cvprblue}{rgb}{0.21,0.49,0.74}
\usepackage[pagebackref,breaklinks,colorlinks,allcolors=cvprblue]{hyperref}

\def\paperID{*****} 
\def\confName{CVPR}
\def\confYear{2025}

\title{Towards classification-based representation learning for place recognition on LiDAR scans}

\author{Maksim Konoplia\\
Moscow Institute of Physics and Technology \\
Moscow, Russia\\
{\tt\small max.konoplia2@gmail.com} \\
\and
Dmitrii Khizbullin\\
King Abdullah University of Science and Technology\\
Thuwal, Saudi Arabia\\
{\tt\small dmitrii.khizbullin@kaust.edu.sa}
}

\begin{document}
\maketitle

\begin{abstract}
Place recognition is a crucial task in autonomous driving, allowing vehicles to determine their position using sensor data. While most existing methods rely on contrastive learning, we explore an alternative approach by framing place recognition as a multi-class classification problem. Our method assigns discrete location labels to LiDAR scans and trains an encoder-decoder model to classify each scan’s position directly. We evaluate this approach on the NuScenes dataset and show that it achieves competitive performance compared to contrastive learning-based methods while offering advantages in training efficiency and stability.

\end{abstract}

\section{Introduction}
\label{sec:intro}

Place recognition using Lidar data is a fundamental task in robotics and autonomous driving, aiming to determine a vehicle's location based on 3D Lidar scans. Accurate localization is essential for navigation, mapping, and decision-making, particularly in environments where GPS signals are unreliable. Over the past decade, significant progress has been made in Lidar-based place recognition, with various approaches leveraging handcrafted descriptors, deep learning-based feature extraction, and cross-modal fusion techniques.

Place recognition can be framed as an instance of information retrieval, where a query scan must be matched against a database of previously recorded scans. This problem shares similarities with face recognition, where embeddings are learned to differentiate between different entities while maintaining invariance to viewpoint and domain shifts. Traditional approaches have focused on contrastive learning-based techniques to generate robust descriptors, but these methods require complex negative sample mining strategies and careful data augmentation.

An alternative to contrastive learning is to formulate place recognition as a multi-class classification problem. However, defining a classification objective is not trivial in this setting, as places do not have rigid class boundaries like objects in standard classification tasks. Instead, a place can have variations in perspective, occlusions, and structural changes over time, making direct class assignment challenging. In this work, we explore how to effectively define and optimize such a classification objective for place recognition. Our primary contribution is a novel formulation of the problem that enables the use of classification-based learning.

\subsection{Early Methods in Place Recognition}

Early advances in place recognition were driven by NetVLAD \cite{Arandjelovi2015NetVLADCA}, which introduced a learnable VLAD-based feature aggregation method for visual localization. PointNet \cite{Qi2016PointNetDL} and PointNet++ \cite{Qi2017PointNetDH} demonstrated the effectiveness of directly processing raw point clouds for 3D tasks. Building on these ideas, PointNetVLAD \cite{Uy2018PointNetVLADDP} extended NetVLAD’s principles to Lidar data by combining PointNet’s permutation-invariant encoding with VLAD-based feature aggregation, enabling large-scale place recognition in unstructured environments. These methods laid the foundation for modern learning-based place recognition, shaping advancements in both visual and Lidar-based localization.

\subsection{Advancements in Lidar-Based Place Recognition}

Recent years have seen significant progress in Lidar-based place recognition, with multiple methods improving the processing of 3D point clouds \cite{Komorowski2020MinkLoc3DPC, Zywanowski2021MinkLoc3DSI3L, Komorowski2022ImprovingPC, Guan2023CrossLoc3DAC}. Among these, the MinkLoc3D family \cite{Komorowski2020MinkLoc3DPC, Zywanowski2021MinkLoc3DSI3L, Komorowski2022ImprovingPC} achieved state-of-the-art performance across several benchmarks. MinkLoc3D employs sparse 3D convolutions within a feature pyramid framework, enhancing feature extraction while maintaining computational efficiency. The integration of generalized-mean pooling further improves descriptor quality, leading to robust place recognition in large-scale environments.

Although convolutional architectures for point cloud-based place recognition had been explored in prior works \cite{Sun2020DAGCED, Liu2018LPDNet3P}, MinkLoc3D demonstrated their effectiveness at scale, reinforcing their importance in modern Lidar-based methods.

\subsection{Multi-Modal Architectures: Image and Point Cloud Fusion}

Integrating multiple sensor modalities, such as camera images and Lidar point clouds, has proven to be an effective strategy for enhancing place recognition performance. This research direction has gained traction in recent years \cite{Zhou2023LCPRAM, Lai2021AdaFusionVF, Pan2023CameraLiDARFW, Cai2024VOLocVP, Puligilla2023LIPLocLI, Komorowski2020MinkLoc3DPC, Xu2023ANP}, leading to notable advancements in the field.

LCPR \cite{Zhou2023LCPRAM} introduced a fusion architecture that combines image and point cloud feature maps at multiple representation levels using self-attention mechanisms. Evaluated on the large-scale NuScenes dataset \cite{Caesar2019nuScenesAM}, LCPR demonstrated superior performance over existing methods. Another notable approach \cite{Xu2023ANP} leveraged image sequences instead of single frames to improve robustness and mitigate outliers. The authors trained their model to learn a similarity function between image sequences and point cloud descriptors, enabling cross-modal retrieval. At inference, the model requires only an image sequence as input, retrieving the corresponding point cloud descriptor for localization, making the method more practical for real-world deployment.

Despite the effectiveness of these approaches, challenges remain in benchmarking cross-modal methods. \cite{Xu2023ANP} highlights the difficulty in direct performance comparison due to varying evaluation setups, as their model's retrieval mechanism differs fundamentally from traditional methods. Further standardized evaluations are necessary to validate the generalization and applicability of such multi-modal architectures in diverse conditions.

\subsection{Recent Advancements in Place Recognition}

A leading approach in modern place recognition is the CrossLoc3D model \cite{Guan2023CrossLoc3DAC}, which addresses the challenge of aligning Lidar descriptors across variations in viewpoint and scan quality. The model enhances robustness by processing voxelized Lidar scans at multiple resolutions, allowing its feature extractor to learn stable structural representations.

A key contribution of \cite{Guan2023CrossLoc3DAC} is the introduction of the CS-Campus3D dataset, an aerial-ground Lidar dataset designed to evaluate place recognition across differing perspectives. The dataset was used for training, validation, and benchmarking, demonstrating that CrossLoc3D outperforms even cross-modal architectures in retrieval accuracy. These results indicate that improving descriptor consistency across viewpoints can significantly enhance place recognition performance, setting a new benchmark in the field.

\subsection{Training Objectives of the Baseline Methods}

PointNetVLAD, LCPR, MinkLoc3D, CrossLoc3D, and OverlapTransformer use the metric learning objective and, specifically, variations of the triplet margin loss for training the embedding model.

\subsection{Datasets}
\label{sec:datasets}

Several publicly available datasets have played a crucial role in advancing place recognition research, particularly in terms of scale and environmental diversity. Among them, KITTI \cite{Geiger2013IJRR}, Oxford RobotCar \cite{Maddern2020RealtimeKG}, NuScenes \cite{Caesar2019nuScenesAM}, and Waymo Open Dataset \cite{Sun_2020_CVPR} stand out as the most widely used.

KITTI, one of the earliest datasets, contains 22 driving sequences with multiple route variations. While limited in scene diversity, it remains a standard benchmark. Oxford RobotCar offers significantly greater variability, featuring over 100 traversals of the same route in Oxford under different weather conditions and times of day, making it ideal for evaluating robustness to environmental changes. NuScenes is the largest and most comprehensive, consisting of 1000 urban driving scenes recorded in Boston and Singapore. It provides high-quality multimodal sensor data across diverse routes and includes an extensive API, facilitating ease of use and integration into research pipelines. The Waymo Open Dataset provides another large-scale alternative, featuring Lidar and camera data collected from multiple locations in the United States.

\subsection{Related works in Face Recognition}

One of the first applications of metric learning \cite{xing2002metriclearning} is face recognition (FR). The early deep face recognition models use the classification objective to train the encoder model: DeepFace \cite{taigman2014deepface}, DeepID \cite{sun2014deeplearningfacerepresentation}. The next generation of FR - using the contrastive objective - was introduced by FaceNet \cite{Schroff_2015} that applied triplet loss instead of classification. However, pure contrastive losses (triplet loss, contrastive loss) were found to be difficult to train efficiently because they require careful sample selection (e.g., hard negative mining) and suffer from slow convergence. The third generation of FR training employed a hybrid approach: returned to the classification objective but enhanced it with contrastive features. The notable examples of these approaches are SphereFace \cite{liu2018sphereface}, CosFace \cite{wang2018cosface}, ArcFace \cite{Deng2018ArcFaceAA}, and AM-Softmax \cite{Wang2018AdditiveMS}. Classification-based training with modified losses proved to provide more stable training and better feature learning, especially for large-scale face datasets. The modified softmax losses encourage compact clusters and large inter-class separability in the feature space.

\subsection{Scalable retrieval with the two-tower architecture}

On the retrieval side, the majority of the works rely on the two-tower architecture. The two-tower architecture, also known as the dual-encoder architecture \cite{huang2013clickthrough,pmlr-v162-menon22a,TwoTowerModelArchitecture,TwoTowerApproachWithExpedia,yan2021system}, allows disentangling the construction of the index and servicing the queries. We follow this architecture in our work as it allows to pre-index the point cloud map and make use of the fast approximate nearest neighbor (ANN) search \cite{malkov2020hnsw} during the query time.

\subsection{Summary of Contributions}

We propose a novel classification-based approach to Lidar-based place recognition, diverging from traditional metric learning methods. To the best of our knowledge, this is the first application of a classification objective for Lidar-based global localization. This formulation allows us to utilize classification loss functions with contrastive enhancements, eliminating the need for complex hard sample mining typically required by contrastive losses. Additionally, we introduce a masked cross-entropy loss to mitigate conflicting gradient updates, ensuring stable encoder training. Finally, we scale our training to a large-scale production setting by integrating data from all available maps, demonstrating the feasibility of our approach in real-world applications.

\section{Method}

\subsection{Problem Statement and Background}

The task of place recognition is to determine the geographic coordinates (either in GPS or a local coordinate system) from which sensor data—such as Lidar scans or images—were acquired. The input data consist of images and point clouds, where a point cloud is a set of 3D coordinates represented as an \( N \times 3 \) matrix, with \( N \) being the number of points in a Lidar scan. A practical solution to this problem must be computationally efficient to support real-time execution, as it is often deployed in autonomous vehicles and robotic systems where latency is critical.

A primary motivation for developing alternative geolocation methods is the unreliability of GPS data. Signal degradation caused by urban canyons, natural obstructions, and adverse weather conditions can significantly impact GPS accuracy, making it insufficient as a standalone localization method. In scenarios of signal loss or low confidence in GPS accuracy, an independent place recognition system is essential.

Another challenge arises from the operational constraints of GPS: for higher localization accuracy, movement is typically required. This dependency affects autonomous vehicle performance, particularly when initiating motion or recovering from system restarts. Without reliable localization, vehicles may face safety risks at critical moments. Consequently, leveraging sensor-based place recognition as a complementary or fallback localization method is crucial for ensuring consistent and robust navigation.

\subsection{Dataset Selection and Analysis}

Among the datasets discussed in \Cref{sec:datasets}, we selected NuScenes due to its scale, data quality, and diversity of sampled locations.

Beyond the number of available samples, a key factor in place recognition datasets is the spatial distribution of recorded locations. A dataset with diverse geographic coverage helps prevent underfitting and improves generalization. The NuScenes dataset provides an effective split for training, validation, and testing, making it well-suited for place recognition. We construct the database using the training split and use the validation and testing splits for query samples. The dataset's temporal segmentation ensures that training and validation samples, while covering overlapping locations, are distinct enough to prevent data leakage and overfitting.

In our research, we have verified that NuScenes samples are distributed along complex, intersecting trajectories. Compared to the single-loop structure of Oxford RobotCar and the limited route repetitions in KITTI, NuScenes provides a more comprehensive spatial distribution for place recognition. Additionally, the dataset includes four distinct geographic areas, further improving diversity and robustness.

\begin{figure}[h]
  \centering
   \includegraphics[width=0.9\linewidth]{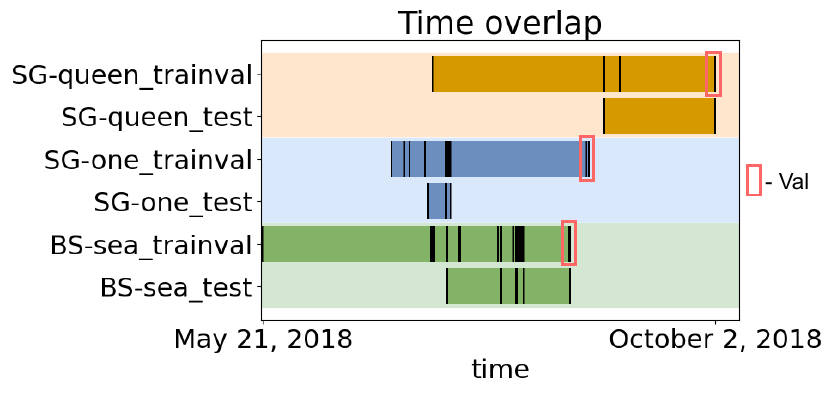}
   \caption{Visualization of time overlap among the four NuScenes maps. Black segments indicate sampling periods, while high-intensity colored segments denote intervals between the first and the last recording. Red bounding boxes indicate samples from validation subset.}
   \label{fig:overlap}
\end{figure}

\subsection{Data Splits}

For place recognition, the dataset must be divided into two primary subsets: the database and the query set. Additionally, the data must be split into training, validation, and test sets for model evaluation.

\subsubsection{Train-val-test split}

The NuScenes dataset provides 850 "trainval" scenes for training and validation, along with 150 scenes for testing.

The key consideration guiding the "trainval" data split into training and validation is that consecutive samples from the same scene should not be assigned to different splits. Since adjacent samples within a scene are highly similar, splitting by individual samples could lead to data leakage. Instead, we sort scenes by timestamp and allocate the first \( k\% \) (where \( k=90 \)) to training, with the remaining scenes assigned to validation. This ensures that validation queries remain sufficiently distinct from training samples.

Figure \ref{fig:overlap} shows NuScenes data samples on the time scale. Two key facts can be derived from the plot. Firstly, the trainval dataset is easily separable into two distinct subsets without overlapping. Secondly, the test set overlaps quite heavily with the trainval data, which led us to the idea to calculate our metrics using validation set. Since the train-val splitting is done considering the necessity for the data samples to come from different scenes shot in different time intervals, metrics calculated on validation subsets are more reliable and valuable.  

\subsubsection{Database-query split}

We construct the database using all training samples, while the validation and testing sets serve as queries. This procedure constitutes in-domain validation. For potential out-of-domain validation, the database should be constructed from a separate sample set, not the training samples. We emphasize that the train-val-test split and the database-query split are two distinct procedures.

An important consideration to keep the recall metrics fair is to make sure that every query sample's discrete location is present in the database. To address this, we follow the filtering strategy used in LCPR \cite{Zhou2023LCPRAM}, removing query samples that lack a corresponding match in the database. Potentially, the unmatched queries can be used to compute the false positive rate or precision of the retrieval; we, however, do not study this.

To define discrete locations, we use a spatial grid rather than relying on continuous coordinates. A grid with cell size \( n \times n \) meters (where \( n=1.0 \) meter) is overlaid on the dataset, and sample coordinates \((x, y)\), where \( x,y \in \mathbb{R} \), are discretized into integer coordinates \((k_{x}, k_{y}, m)\), where \( k_{x}, k_{y} \in \mathbb{Z} \) and \( m \) represents a unique map identifier (ranging from 0 to 3 by the number of maps in NuScenes).

Table \ref{tab:data_organisation} presents dataset statistics. Training samples form the database, while validation and test samples serve as queries. \( N \) denotes the total number of samples before filtering, while \( N' \) indicates the number of remaining samples after filtering unmatched locations. $C_{DB}$ stands for the number of classes present in database (train) and $C_q$ stands for the query classes (validation and testing).

\begin{table}[ht]
\centering
\caption{Quantities of samples for train-val-test splits on all NuScenes maps.}
\label{tab:data_organisation}
\begin{tabular}{|c|cccc|}
\textbf{Map} & Boston & SG-one & SG-holland & SQ-queens \\ \hline
\textbf{$N$ Train} & 16906 & 6577 & 3084 & 4166\\ \hline
\textbf{$N$ Val} & 1879 & 731 & 343 & 463\\ \hline
\textbf{$N^{`}$ Val} & 1229 & 478 & 224 & 302 \\ \hline
\textbf{$N$ Test} & 3318 & 796 & - & 1894\\ \hline
\textbf{$N^{`}$ Test} & 1327 & 318 & - & 757 \\ \hline
\textbf{$C_{DB}$} & 9086 & 4536 & 691 & 2748 \\ \hline
\textbf{$C_q$ Val} & 1763 & 274 & 81 & 117 \\ \hline
\textbf{$C_q$ Test} & 1373 & 129 & - & 315 \\ \hline
\end{tabular}
\end{table}

\subsection{ML problem statement and annotation structure}

While all recent papers we came across solve Place Recognition as a contrastive learning problem, we formulate it as a multi-class classification one. We build a common classification space that includes all the maps in the training set combined.

\subsubsection{Discrete Location Notation for Multiple Maps}

Assume we have several maps, each identified by a unique map index. For a given map indexed by \(m\), let the samples be denoted by
\[
p_{i}^{(m)} = \bigl(x_{i}^{(m)}, y_{i}^{(m)}, z_{i}^{(m)}\bigr), \quad \text{with } z_{i}^{(m)} = 0,
\]
where \(i = 1, 2, \dots, N_m\) and \(N_m\) is the number of samples in map \(m\).

We define the discretized coordinates as follows:
\[
k_x = \left\lfloor \frac{x}{h} \right\rfloor \quad \text{and} \quad k_y = \left\lfloor \frac{y}{h} \right\rfloor,
\]
where \(h\) is the grid size that defines the size of each grid cell. For each sample in map \(m\), we then define a discrete grid index by:
\[
\mathbf{g}_{i}^{(m)} = \bigl(k_{x,i}^{(m)}, k_{y,i}^{(m)}, m\bigr),
\]
with 
\[
k_{x,i}^{(m)} = \left\lfloor \frac{x_{i}^{(m)}}{h} \right\rfloor \quad \text{and} \quad k_{y,i}^{(m)} = \left\lfloor \frac{y_{i}^{(m)}}{h} \right\rfloor.
\]
Here, the first two components represent the discretized \(x\) and \(y\) coordinates, while the third component indicates the map to which the sample belongs.

\begin{itemize}
    \item \(\mathbf{g}_{i}^{(m)}\): The discretized grid index for the \(i\)-th sample in map \(m\).
    \item \(m\): The map index, uniquely identifying each map.
    \item \(i\): An index running over all samples within a given map, i.e., \(i = 1, 2, \dots, N_m\).
    \item \(x_{i}^{(m)}\) and \(y_{i}^{(m)}\): The original continuous \(x\) and \(y\) coordinates of the \(i\)-th sample in map \(m\).
    \item \(h\): The grid size, defining the size of each grid cell.
\end{itemize}

\subsubsection{Class Label Construction}

We construct class labels by assigning a continuous numbering to each unique 3-tuple \(\mathbf{g} = (k_x, k_y, m)\) (with \(k_x\) and \(k_y\) as defined above). This is achieved by first ordering all possible 3-tuples lexicographically. Each unique 3-tuple is then assigned a unique class label from the set \(\{0, 1, \dots, C-1\}\), where \(C\) is the total number of distinct 3-tuples across all maps. In this way, the mapping
\[
\phi: (k_x, k_y, m) \mapsto \{0, 1, \dots, C-1\}
\]
provides a continuous numbering that is used as class labels.

The total number of unique classes annotated in the result of the described method for the NuScenes dataset is 17061.

\begin{figure}[h]
  \centering
   \includegraphics[width=0.9\linewidth]{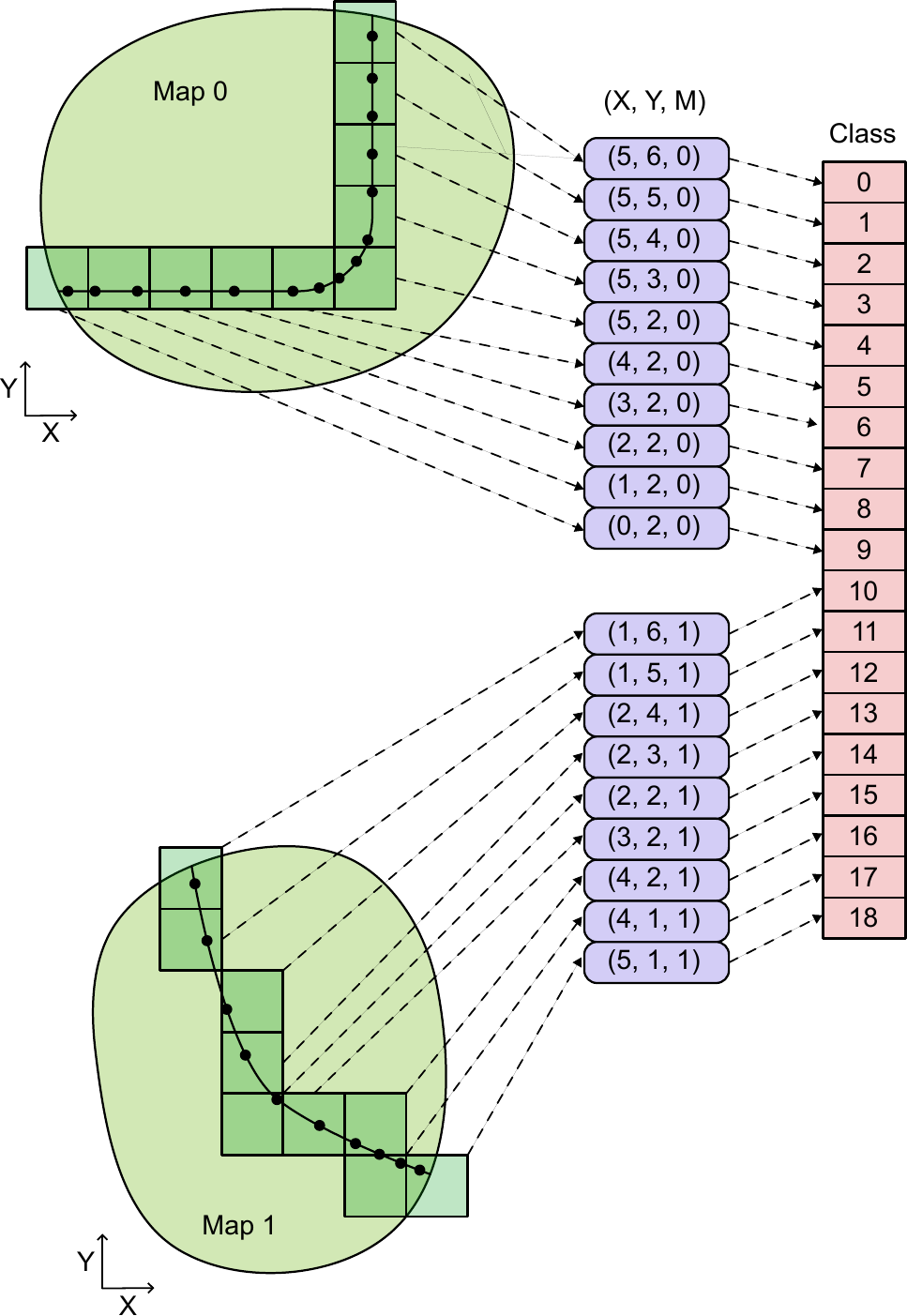}
   \caption{Proposed method for the formation of class labels for location recognition. Black circles are locations where the scans were taken, the solid curved lines are the trajectory of the vehicle. Dark green boxes are non-empty cells. The 3-tuples are formed as X and Y integer coordinates of cells inside a map, extended with the map identifier M. The class labels are formed as a continuous numbering of all 3-tuples.}
   \label{fig:labels}
\end{figure}

See \Cref{fig:labels} for the illustration of the label construction procedure.

\subsection{Neural Network Architecture}

Our model consists of an encoder and a decoder. The encoder employs PointNet++ \cite{Qi2017PointNetDH} as a backbone for processing point clouds and an embedding fully-connected layer that forms the embeddings of size 512. The decoder is a single fully-connected layer that converts embeddings into class probabilities. The class probabilities are then used to compute the Masked Cross-Entropy loss (see Section \ref{Training setup}) with respect to ground-truth location labels.

\begin{figure*}[h]
  \centering
   \includegraphics[width=0.9\linewidth]{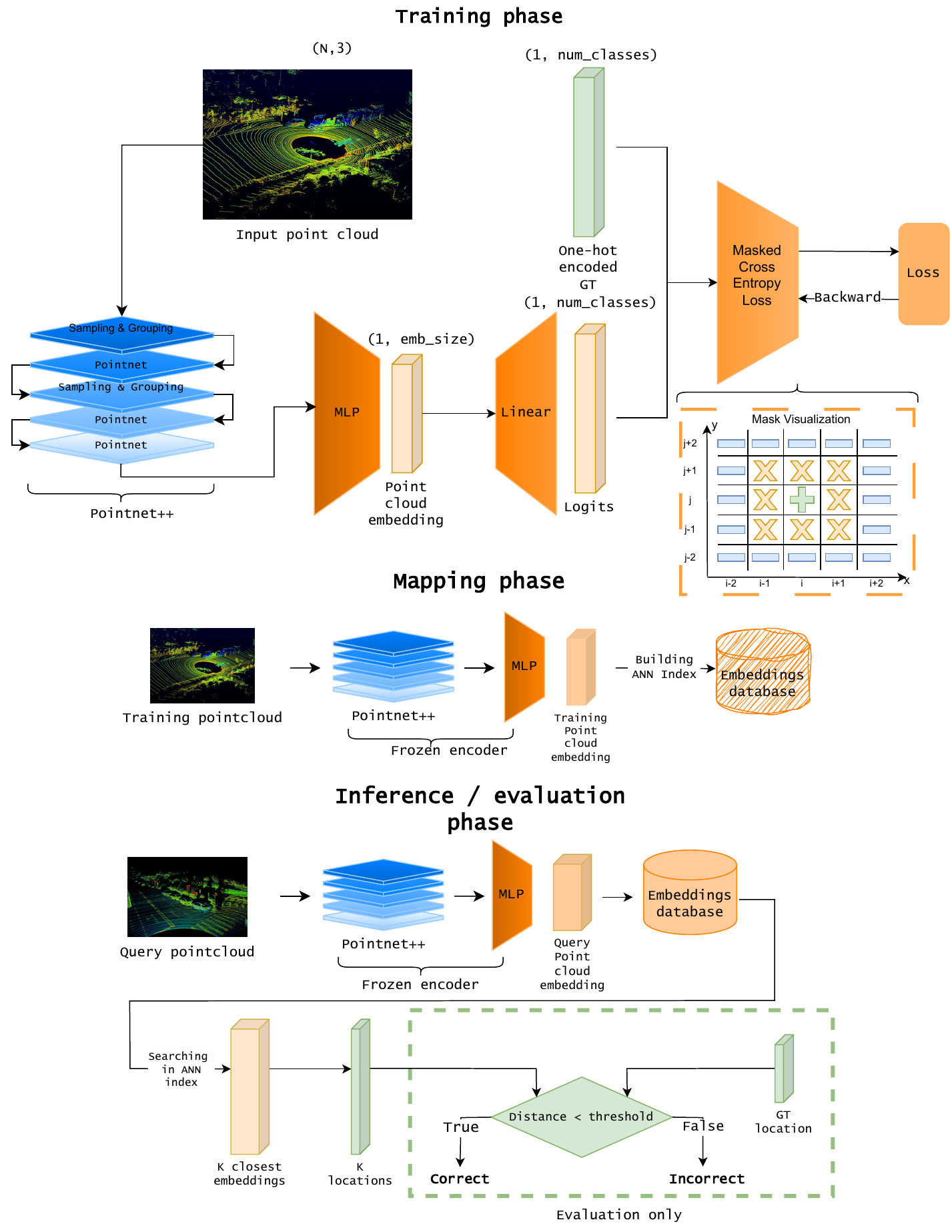}
   \caption{Model architecture: Training phase. The model consists of a PointNet++ backbone, a linear embedding layer, and a linear classification head that outputs class probabilities. Masked cross-entropy loss is used to train the network (see Section \ref{Training setup}). Inference phase: The frozen backbone extracts embeddings from the query point cloud, which are then used to perform a KNN search in the pre-indexed database to retrieve the closest matches.}
   \label{fig:arch_merged}
\end{figure*}

During validation, all training database embeddings are computed and saved to a k-nearest neighbors (KNN) index. Then, embeddings are computed for every query point cloud. A KNN search is performed within the collected embedding database to retrieve the nearest neighbors. The model is considered to make a correct prediction if the embedding corresponding to the correct query location appears among the \( K \) nearest neighbors. The overall architecture is illustrated in Figures \ref{fig:arch_merged} and \ref{fig:arch_merged}.

\subsection{Training and Evaluation Procedures}

\subsubsection{Training Setup}
\label{Training setup}

The model is trained as a classification task, with the objective that the encoder-generated embeddings become highly informative and discriminative. During training, the class probabilities produced by the decoder are used to compute the masked cross-entropy loss.

To prevent penalizing predictions that fall within adjacent cells to the correct location, these class probabilities are ignored in the loss calculation by assigning them a value of \(-\infty\), as illustrated in \Cref{fig:arch_merged}. Green plus sign represent the positive class, blue minus signs indicate negative classes, and orange crosses denote ignored cells.

\subsubsection{Evaluation Setup}

In this section, we describe the in-domain evaluation procedure, meaning that the database is created from the training samples, not from the samples of a hold-out map, which would be out-of-domain evaluation.

During evaluation, the classification head is removed, and only the encoder is used to generate embeddings. These embeddings are computed for all training samples and stored in an approximate nearest neighbor (ANN) index, forming the database. Query samples are then embedded using the same encoder, and a KNN search is performed within the database. The retrieved neighbors' class labels are mapped back to their original coordinates using $\phi^{-1}$, allowing the computation of the actual spatial distance between the predicted and ground-truth positions. A prediction is considered correct if the distance falls within a threshold of 18 meters following LCPR. Even though the database is created for the entire training set of samples from all maps, during the query phase, for a query sample belonging to map $m$, we search only among the database samples that belong to map $m$. We perform evaluation on the queries from the validation set, while the test set queries are not used.

The primary evaluation metric is Mean Average Recall (MAR), reported for both top-1 and top-1\% retrieved results. The top-1\% threshold is determined as 1\% of the total number of classes, which is 170 in this setup.

\section{Results}

\label{Experimental results}
\subsection{Experimental results}

We conducted a series of experiments measuring average Recall@1 and Recall@1\% on 4 different NuScence maps: Boston, Singapore One North, Singapore Holland Village and Singapore Queenstown. The same model was used for all the maps, that was trained on entire training dataset. We came to meaningful conclusions regarding some hyper parameters of the model and augmentation and regularization techniques used.

One of the key findings were the benefits of using point decimation augmentation. This transformation leaves only 1/k part of points in the point cloud randomly, where k=1, 5, 10, 20 is a parameter. This highly reduces the time and memory needed for the training process and vastly increases robustness of the model, which results in better metrics. Another finding of the conducted experiments was the positive effect of embedding normalization.

Experiment results are depicted in Table \ref{tab:main_experiments}. The radius of match was set to 18 meters, and the batch size was set to 64.

\begin{table}[ht]
\centering
\caption{Performance comparison with foundation models and LCPR on the NuScenes dataset. The first row represents \textbf{Recall @ 1}, with results for \textbf{Boston} and \textbf{SG-one} in the second row. Metrics are sourced from Table 2 of \cite{Zhou2023LCPRAM}.}
\label{tab:perf}
\resizebox{\linewidth}{!}{%
\begin{tabular}{lcc}
\hline
\textbf{Model} & \multicolumn{2}{c}{\textbf{Recall @ 1}} \\
 & \textbf{Boston} & \textbf{SG-one} \\
\hline
\textbf{PointNetVLAD} \cite{Uy2018PointNetVLADDP} & 74.30 & 98.49 \\
\textbf{OverlapTransformer} \cite{Ma2022OverlapTransformerAE} & 82.05 & 98.73 \\
\textbf{LCPR-L} \cite{Zhou2023LCPRAM} & 64.69 & -- \\
\textbf{Ours} & 39.13 & 56.34 \\
\hline
\end{tabular}%
}
\end{table}

The current state of the approach does not seem to keep up with the models trained via contrastive learning. One of the implications would be that the feature extractor we are using is too lightweight and does not provide enough learning capacity for classification loss to work with. One of the supports would be the fact that LCPR model has three times as much parameters (29,97 M) as our best model (10.3M), and the authors of LCPR claims that their model has less parameters than any of the models they compare their solution with.

\section{Discussion}

\subsection{Comparison with LCPR}

LCPR shows higher score on its LiDAR-only version (LCPR-L). However, LCPR train a single model just using data sample from Boston map, and test on the same map, effectively reporting training metrics of models highly overfitted to the specific map. We, however, train a single model on all maps combined which makes our model generalize better. Still our evaluation protocol follows LCPR, as we do not perform the most fair variant of evaluation either: out-of-domain evaluation on a map that was never seen during the training phase.

\subsection{Out-of-domain evaluation}

The most reliable way to assess the quality of the embedding model is to test it in the out-of-domain setting. One option to do it is to apply leave-one-out pattern as follows. Given the 4 available maps, create 4 datasets where 3 maps go into the training set, and the remaining map goes into the validation; train 4 models and validate them on the corresponding hold-out map. For a hold-out-map, the database set of samples (LiDAR scans) and the query set of samples should be created independently from the union of the three training maps. In this manner, each trained model will be evaluated on the unseen data. The goal of this "hold-one-map-out" procedure is to get fair quality metrics. During the application / inference operation of the model, either an ensemble of the 4 trained models can be applied with yet to be researched kind of aggregation of the embedding vectors (average, concatenation) or the final model can be trained blindly on all 4 maps without validation. We leave this to future work.

\subsection{Density of annotation and scaling of the per-cell classification.}

The number of classes, around 17 thousand on all nuScenes maps, is not high compared to large-scale face recognition datasets with over 1 million identities, since the driving trajectory is basically 1-dimensional - a curved line in 2D space. If we had more dense coverage by LiDAR scans then, for  example, for all 1 m x 1 m cells of 1 km x 1 km area, we would get as many as 1 million cells/classes. For areas on hte scale of a single city, it would be 10s-100s of millions classes. We do not study the scaling of our solution to this level of complexity. Potential sources of obstacles are out-of-memory (OOM) situation in the decoder, as well as from the gradients of negative classes may hypothetically overrun the gradients from one positive class.

Potentially, the multi-class formulation can be efficient for localization based on satellite imaging, where the data intrinsically contains features for all locations/cells. For satellite images, training with the contrastive/triplet formulation may be inefficient and or intangible since the possible number of pairs grows as $O(N^2)$ from the number of cells and as $O(N^4)$ from the linear size of the area (its "diameter"). We hypothesize that training a location recognition model for satellite (and any other location-wise dense 2D) data is more efficient using the classification setting, even though the OOM limitation will most likely have to be addressed.

\subsection{The effect of the radius of the match}

The evaluation protocol that is currently adopted in the literature uses the radius of match of 18 meters to positively assess the closeness of the predicted location to the ground truth location. For practical application, 18 meters may be an extremely rough estimate. Furthermore, the scores that, for example, LCPR achieves for such rough matches are over 99\% top-1, which are already non-discriminative and give the false impression of the problem having been solved. We advocate that the radius of the match should be reduced to 2 meters. Out classification-based method has a potential to thrive in these conditions since it makes the network learn from hard samples more intensively by classifying all spatially correlated samples as negative class. 

\subsection{Relation to the sizes of FR datasets}

To compare the complexity of classification-based location recognition with face recognition, we collect the numbers of identities of the popular FR datasets in \cref{tab:fr_datasets}. We can see that the scaling potential of classification-based location recognition training is more than two orders of magnitude based on the comparison to face recognition.

\begin{table}[ht]
\centering
\caption{FR datasets and their number of identities.}
\label{tab:fr_datasets}
\resizebox{\linewidth}{!}{%
\begin{tabular}{l r}
\multicolumn{2}{c}{Face recognition} \\
\hline
\textbf{Dataset} & \textbf{\# Identities} \\
\hline
Labeled Faces in the Wild (LFW) \cite{huang2008lfw} & 5,749 \\
CelebFaces Attributes (CelebA) \cite{liu2015faceattributes} & 10,177 \\
MegaFace \cite{miller2015megaface} & 672,057 \\
MS-Celeb-1M \cite{guo2016msceleb1m} & 100,000 \\
WebFace260M \cite{zhu2022webface260m} & 4,000,000 \\
\hline
\\
\multicolumn{2}{c}{Location recognition} \\
\hline
\textbf{Method, Dataset} & \textbf{\# Descrete Locations} \\
\hline
Ours, NuScenes & 17,061 \\
\hline
\end{tabular}%
}
\end{table}

\subsection{Conclusion}

Our work shows that training a point cloud embedding via the classification problem formulation is a viable approach for place recognition. The reported scores of our solution are lower than the state-of-the-art methods; nevertheless, we've shown that this approach is feasible in principle.

\subsection{Limitations}

All the evaluations were performed in the in-domain setting; therefore, the applicability of our approach for extracting embeddings from point clouds acquired outside the training data set has not yet been verified.

One of the biggest possible risks of applying the classification approach is scaling w.r.t the quantity locations. The 4 maps of NuScenes produce 17k classes. However, for much bigger maps, number of 1 m x 1 m locations can count in millions or more, which will make the decoder matrix (embedding size times the number of classes) impossible to store in the RAM.

The effect of the choice of the cell size (1 m x 1 m in our experiments) was not studied.

{
    \small
    \bibliographystyle{ieeenat_fullname}
    \bibliography{main}

\begin{thebibliography}{40}
\providecommand{\natexlab}[1]{#1}
\providecommand{\url}[1]{\texttt{#1}}
\expandafter\ifx\csname urlstyle\endcsname\relax
  \providecommand{\doi}[1]{doi: #1}\else
  \providecommand{\doi}{doi: \begingroup \urlstyle{rm}\Url}\fi

\bibitem[Arandjelovi{\'c} et~al.(2015)Arandjelovi{\'c}, Gron{\'a}t, Torii, Pajdla, and Sivic]{Arandjelovi2015NetVLADCA}
Relja Arandjelovi{\'c}, Petr Gron{\'a}t, Akihiko Torii, Tom{\'a}s Pajdla, and Josef Sivic.
\newblock Netvlad: Cnn architecture for weakly supervised place recognition.
\newblock \emph{2016 IEEE Conference on Computer Vision and Pattern Recognition (CVPR)}, pages 5297--5307, 2015.

\bibitem[Caesar et~al.(2019)Caesar, Bankiti, Lang, Vora, Liong, Xu, Krishnan, Pan, Baldan, and Beijbom]{Caesar2019nuScenesAM}
Holger Caesar, Varun Bankiti, Alex~H. Lang, Sourabh Vora, Venice~Erin Liong, Qiang Xu, Anush Krishnan, Yuxin Pan, Giancarlo Baldan, and Oscar Beijbom.
\newblock {nuScenes}: A multimodal dataset for autonomous driving.
\newblock \emph{2020 IEEE/CVF Conference on Computer Vision and Pattern Recognition (CVPR)}, pages 11618--11628, 2019.

\bibitem[Cai et~al.(2024)Cai, Wang, zhe Huang, Shao, and Li]{Cai2024VOLocVP}
Xudong Cai, Yongcai Wang, zhe Huang, Yu Shao, and Deying Li.
\newblock Voloc: Visual place recognition by querying compressed lidar map.
\newblock \emph{2024 IEEE International Conference on Robotics and Automation (ICRA)}, pages 10192--10199, 2024.

\bibitem[Deng et~al.(2018)Deng, Guo, and Zafeiriou]{Deng2018ArcFaceAA}
Jiankang Deng, J. Guo, and Stefanos Zafeiriou.
\newblock Arcface: Additive angular margin loss for deep face recognition.
\newblock \emph{2019 IEEE/CVF Conference on Computer Vision and Pattern Recognition (CVPR)}, pages 4685--4694, 2018.

\bibitem[Geiger et~al.(2013)Geiger, Lenz, Stiller, and Urtasun]{Geiger2013IJRR}
Andreas Geiger, Philip Lenz, Christoph Stiller, and Raquel Urtasun.
\newblock Vision meets robotics: The kitti dataset.
\newblock \emph{International Journal of Robotics Research (IJRR)}, 2013.

\bibitem[Guan et~al.(2023)Guan, Muthuselvam, Hoover, Wang, Liang, Sathyamoorthy, Conover, and Manocha]{Guan2023CrossLoc3DAC}
Tianrui Guan, Aswath Muthuselvam, Montana Hoover, Xijun Wang, Jing Liang, Adarsh~Jagan Sathyamoorthy, Damon~M. Conover, and Dinesh Manocha.
\newblock Crossloc3d: Aerial-ground cross-source 3d place recognition.
\newblock \emph{2023 IEEE/CVF International Conference on Computer Vision (ICCV)}, pages 11301--11310, 2023.

\bibitem[Guo et~al.(2016)Guo, Zhang, Hu, He, and Gao]{guo2016msceleb1m}
Yandong Guo, Lei Zhang, Yuxiao Hu, Xiaodong He, and Jianfeng Gao.
\newblock Ms-celeb-1m: A dataset and benchmark for large-scale face recognition, 2016.

\bibitem[Huang et~al.(2008)Huang, Mattar, Berg, and Learned-Miller]{huang2008lfw}
Gary Huang, Marwan Mattar, Tamara Berg, and Eric Learned-Miller.
\newblock Labeled faces in the wild: A database forstudying face recognition in unconstrained environments.
\newblock \emph{Tech. rep.}, 2008.

\bibitem[Huang et~al.(2013)Huang, He, Gao, Deng, Acero, and Heck]{huang2013clickthrough}
Po-Sen Huang, Xiaodong He, Jianfeng Gao, Li Deng, Alex Acero, and Larry Heck.
\newblock Learning deep structured semantic models for web search using clickthrough data.
\newblock In \emph{Proceedings of the 22nd ACM International Conference on Information \& Knowledge Management}, page 2333–2338, New York, NY, USA, 2013. Association for Computing Machinery.

\bibitem[Komorowski(2020)]{Komorowski2020MinkLoc3DPC}
Jacek Komorowski.
\newblock Minkloc3d: Point cloud based large-scale place recognition.
\newblock \emph{2021 IEEE Winter Conference on Applications of Computer Vision (WACV)}, pages 1789--1798, 2020.

\bibitem[Komorowski(2022)]{Komorowski2022ImprovingPC}
Jacek Komorowski.
\newblock Improving point cloud based place recognition with ranking-based loss and large batch training.
\newblock \emph{2022 26th International Conference on Pattern Recognition (ICPR)}, pages 3699--3705, 2022.

\bibitem[Kumar(2024)]{TwoTowerModelArchitecture}
Sumit Kumar.
\newblock Two tower model architecture: Current state and promising extensions, 2024.

\bibitem[Lai et~al.(2021)Lai, Yin, and Scherer]{Lai2021AdaFusionVF}
Haowen Lai, Peng Yin, and Sebastian~A. Scherer.
\newblock Adafusion: Visual-lidar fusion with adaptive weights for place recognition.
\newblock \emph{IEEE Robotics and Automation Letters}, 7:\penalty0 12038--12045, 2021.

\bibitem[Liu et~al.(2018{\natexlab{a}})Liu, Wen, Yu, Li, Raj, and Song]{liu2018sphereface}
Weiyang Liu, Yandong Wen, Zhiding Yu, Ming Li, Bhiksha Raj, and Le Song.
\newblock Sphereface: Deep hypersphere embedding for face recognition, 2018{\natexlab{a}}.

\bibitem[Liu et~al.(2015)Liu, Luo, Wang, and Tang]{liu2015faceattributes}
Ziwei Liu, Ping Luo, Xiaogang Wang, and Xiaoou Tang.
\newblock Deep learning face attributes in the wild.
\newblock In \emph{Proceedings of International Conference on Computer Vision (ICCV)}, 2015.

\bibitem[Liu et~al.(2018{\natexlab{b}})Liu, Zhou, Suo, Yin, Chen, Wang, Li, and Liu]{Liu2018LPDNet3P}
Zhe Liu, Shunbo Zhou, Chuanzhe Suo, Peng Yin, Wen Chen, Hesheng Wang, Haoang Li, and Yunhui Liu.
\newblock Lpd-net: 3d point cloud learning for large-scale place recognition and environment analysis.
\newblock \emph{2019 IEEE/CVF International Conference on Computer Vision (ICCV)}, pages 2831--2840, 2018{\natexlab{b}}.

\bibitem[Ma et~al.(2022)Ma, Zhang, Xu, Ai, Gu, and Chen]{Ma2022OverlapTransformerAE}
Junyi Ma, Jun Zhang, Jintao Xu, Rui Ai, Weihao Gu, and Xieyuanli Chen.
\newblock Overlaptransformer: An efficient and yaw-angle-invariant transformer network for lidar-based place recognition.
\newblock \emph{IEEE Robotics and Automation Letters}, 7:\penalty0 6958--6965, 2022.

\bibitem[Maddern et~al.(2020)Maddern, Pascoe, Gadd, Barnes, Yeomans, and Newman]{Maddern2020RealtimeKG}
William~P. Maddern, Geoffrey Pascoe, Matthew Gadd, Dan Barnes, Brian Yeomans, and Paul Newman.
\newblock Real-time kinematic ground truth for the oxford robotcar dataset.
\newblock \emph{ArXiv}, abs/2002.10152, 2020.

\bibitem[Malkov and Yashunin(2020)]{malkov2020hnsw}
Yu~A. Malkov and D.~A. Yashunin.
\newblock Efficient and robust approximate nearest neighbor search using hierarchical navigable small world graphs.
\newblock \emph{IEEE Trans. Pattern Anal. Mach. Intell.}, 42\penalty0 (4):\penalty0 824–836, 2020.

\bibitem[Menon et~al.(2022)Menon, Jayasumana, Rawat, Kim, Reddi, and Kumar]{pmlr-v162-menon22a}
Aditya Menon, Sadeep Jayasumana, Ankit~Singh Rawat, Seungyeon Kim, Sashank Reddi, and Sanjiv Kumar.
\newblock In defense of dual-encoders for neural ranking.
\newblock In \emph{Proceedings of the 39th International Conference on Machine Learning}, pages 15376--15400. PMLR, 2022.

\bibitem[Miller et~al.(2015)Miller, Brossard, Seitz, and Kemelmacher-Shlizerman]{miller2015megaface}
D. Miller, E. Brossard, S. Seitz, and I. Kemelmacher-Shlizerman.
\newblock Megaface: A million faces for recognition at scale, 2015.

\bibitem[Pan et~al.(2023)Pan, Xie, Wu, and Zhou]{Pan2023CameraLiDARFW}
Yan Pan, Jiapeng Xie, Jiajie Wu, and Bo Zhou.
\newblock Camera-lidar fusion with latent contact for place recognition in challenging cross-scenes.
\newblock \emph{ArXiv}, abs/2310.10371, 2023.

\bibitem[Puligilla et~al.(2023)Puligilla, Omama, Zaidi, Parihar, and Krishna]{Puligilla2023LIPLocLI}
Sai~Shubodh Puligilla, Mohammad Omama, Husain Zaidi, Udit~Singh Parihar, and Madhava Krishna.
\newblock Lip-loc: Lidar image pretraining for cross-modal localization.
\newblock \emph{2024 IEEE/CVF Winter Conference on Applications of Computer Vision Workshops (WACVW)}, pages 939--948, 2023.

\bibitem[Qi et~al.(2016)Qi, Su, Mo, and Guibas]{Qi2016PointNetDL}
C. Qi, Hao Su, Kaichun Mo, and Leonidas~J. Guibas.
\newblock Pointnet: Deep learning on point sets for 3d classification and segmentation.
\newblock \emph{2017 IEEE Conference on Computer Vision and Pattern Recognition (CVPR)}, pages 77--85, 2016.

\bibitem[Qi et~al.(2017)Qi, Yi, Su, and Guibas]{Qi2017PointNetDH}
C. Qi, L. Yi, Hao Su, and Leonidas~J. Guibas.
\newblock Pointnet++: Deep hierarchical feature learning on point sets in a metric space.
\newblock \emph{ArXiv}, abs/1706.02413, 2017.

\bibitem[Rincon(2023)]{TwoTowerApproachWithExpedia}
Eric Rincon.
\newblock Candidate generation using a two tower approach with expedia group traveler data, 2023.

\bibitem[Schroff et~al.(2015)Schroff, Kalenichenko, and Philbin]{Schroff_2015}
Florian Schroff, Dmitry Kalenichenko, and James Philbin.
\newblock Facenet: A unified embedding for face recognition and clustering.
\newblock In \emph{2015 IEEE Conference on Computer Vision and Pattern Recognition (CVPR)}, page 815–823. IEEE, 2015.

\bibitem[Sun et~al.(2020{\natexlab{a}})Sun, Kretzschmar, Dotiwalla, Chouard, Patnaik, Tsui, Guo, Zhou, Chai, Caine, Vasudevan, Han, Ngiam, Zhao, Timofeev, Ettinger, Krivokon, Gao, Joshi, Zhang, Shlens, Chen, and Anguelov]{Sun_2020_CVPR}
Pei Sun, Henrik Kretzschmar, Xerxes Dotiwalla, Aurelien Chouard, Vijaysai Patnaik, Paul Tsui, James Guo, Yin Zhou, Yuning Chai, Benjamin Caine, Vijay Vasudevan, Wei Han, Jiquan Ngiam, Hang Zhao, Aleksei Timofeev, Scott Ettinger, Maxim Krivokon, Amy Gao, Aditya Joshi, Yu Zhang, Jonathon Shlens, Zhifeng Chen, and Dragomir Anguelov.
\newblock Scalability in perception for autonomous driving: Waymo open dataset.
\newblock In \emph{Proceedings of the IEEE/CVF Conference on Computer Vision and Pattern Recognition (CVPR)}, 2020{\natexlab{a}}.

\bibitem[Sun et~al.(2020{\natexlab{b}})Sun, Liu, He, Fan, and Du]{Sun2020DAGCED}
Qi Sun, Hongyan Liu, Jun He, Zhaoxin Fan, and Xiaoyong Du.
\newblock Dagc: Employing dual attention and graph convolution for point cloud based place recognition.
\newblock \emph{Proceedings of the 2020 International Conference on Multimedia Retrieval}, 2020{\natexlab{b}}.

\bibitem[Sun et~al.(2014)Sun, Wang, and Tang]{sun2014deeplearningfacerepresentation}
Yi Sun, Xiaogang Wang, and Xiaoou Tang.
\newblock Deep learning face representation by joint identification-verification, 2014.

\bibitem[Taigman et~al.(2014)Taigman, Yang, Ranzato, and Wolf]{taigman2014deepface}
Yaniv Taigman, Ming Yang, Marc'Aurelio Ranzato, and Lior Wolf.
\newblock Deepface: Closing the gap to human-level performance in face verification.
\newblock 2014.

\bibitem[Uy and Lee(2018)]{Uy2018PointNetVLADDP}
Mikaela~Angelina Uy and Gim~Hee Lee.
\newblock Pointnetvlad: Deep point cloud based retrieval for large-scale place recognition.
\newblock \emph{2018 IEEE/CVF Conference on Computer Vision and Pattern Recognition}, pages 4470--4479, 2018.

\bibitem[Wang et~al.(2018{\natexlab{a}})Wang, Cheng, Liu, and Liu]{Wang2018AdditiveMS}
Feng Wang, Jian Cheng, Weiyang Liu, and Haijun Liu.
\newblock Additive margin softmax for face verification.
\newblock \emph{IEEE Signal Processing Letters}, 25:\penalty0 926--930, 2018{\natexlab{a}}.

\bibitem[Wang et~al.(2018{\natexlab{b}})Wang, Wang, Zhou, Ji, Gong, Zhou, Li, and Liu]{wang2018cosface}
Hao Wang, Yitong Wang, Zheng Zhou, Xing Ji, Dihong Gong, Jingchao Zhou, Zhifeng Li, and Wei Liu.
\newblock Cosface: Large margin cosine loss for deep face recognition, 2018{\natexlab{b}}.

\bibitem[Xing et~al.(2002)Xing, Jordan, Russell, and Ng]{xing2002metriclearning}
Eric Xing, Michael Jordan, Stuart~J Russell, and Andrew Ng.
\newblock Distance metric learning with application to clustering with side-information.
\newblock In \emph{Advances in Neural Information Processing Systems}. MIT Press, 2002.

\bibitem[Xu et~al.(2023)Xu, Liu, Meng, and Sun]{Xu2023ANP}
Huaiyuan Xu, Huaping Liu, Shiyu Meng, and Yuxiang Sun.
\newblock A novel place recognition network using visual sequences and lidar point clouds for autonomous vehicles.
\newblock \emph{2023 IEEE 26th International Conference on Intelligent Transportation Systems (ITSC)}, pages 2862--2867, 2023.

\bibitem[Yan(2021)]{yan2021system}
Ziyou Yan.
\newblock System design for recommendations and search.
\newblock \emph{eugeneyan.com}, 2021.

\bibitem[Zhou et~al.(2023)Zhou, Xu, Xiong, and Ma]{Zhou2023LCPRAM}
Zijie Zhou, Jingyi Xu, Guangming Xiong, and Junyi Ma.
\newblock Lcpr: A multi-scale attention-based lidar-camera fusion network for place recognition.
\newblock \emph{IEEE Robotics and Automation Letters}, 9:\penalty0 1342--1349, 2023.

\bibitem[Zhu et~al.(2022)Zhu, Huang, Deng, Ye, Huang, Chen, Zhu, Yang, Du, Lu, and Zhou]{zhu2022webface260m}
Zheng Zhu, Guan Huang, Jiankang Deng, Yun Ye, Junjie Huang, Xinze Chen, Jiagang Zhu, Tian Yang, Dalong Du, Jiwen Lu, and Jie Zhou.
\newblock Webface260m: A benchmark for million-scale deep face recognition, 2022.

\bibitem[Zywanowski et~al.(2021)Zywanowski, Banaszczyk, Nowicki, and Komorowski]{Zywanowski2021MinkLoc3DSI3L}
Kamil Zywanowski, Adam Banaszczyk, Michał~R. Nowicki, and Jacek Komorowski.
\newblock Minkloc3d-si: 3d lidar place recognition with sparse convolutions, spherical coordinates, and intensity.
\newblock \emph{IEEE Robotics and Automation Letters}, PP:\penalty0 1--1, 2021.

\end{thebibliography}
}


\appendix

\section{Hyperparameter sweep}

We collect results of the experiments with various hyperparameter configurations in \Cref{tab:main_experiments}.

\begin{table*}[!htb]
\centering
\caption{Experiments across hyperparameters.}
\label{tab:main_experiments}
{ 
\begin{tabular}{|c|c|c|c|c|c|c|c|c|c|c|}
\hline
\textbf{Emb} & \textbf{Point} & \textbf{Emb} & \multirow{2}{*}{\textbf{Metric}} & \textbf{Boston} & \textbf{SG} & \textbf{SG} & \textbf{SG} & \textbf{Epoch} & \textbf{Train} & \textbf{Param} \\
\textbf{Size} & \textbf{Decim} & \textbf{Norm} & & & \textbf{One North} & \textbf{Queen} & \textbf{Holland} & \textbf{Time (s)} & \textbf{Time (h)} & \\ \hline
16   & 20 & off & Recall@1    & 0.30 & 0.41 & 0.55 & 0.58 & 40  & 6h & 1.8M \\ \cline{4-8}
\rowcolor{LightGray}
      &    &     & Recall@1\%  & 0.92 & 0.89 & 0.98 & 1    &      &  &   \\ \hline
512  & 20 & off & Recall@1    & 0.39 & 0.53 & 0.82 & 0.88 & 40  & 7h & 10.3M \\ \cline{4-8}
\rowcolor{LightGray}
      &    &     & Recall@1\%  & 0.93 & 0.96 & 1    & 1    &      &  &   \\ \hline
1024 & 20 & off & Recall@1    & 0.38 & 0.51 & 0.79 & 0.93 & 40  & 8h & 19.2M \\ \cline{4-8}
\rowcolor{LightGray}
      &    &     & Recall@1\%  & 0.92 & 0.95 & 1    & 1    &      &   &  \\ \hline
2048 & 20 & off & Recall@1    & 0.36 & 0.53 & 0.81 & 0.86 & 41  & 8h & 36.9M \\ \cline{4-8}
\rowcolor{LightGray}
      &    &     & Recall@1\%  & 0.91 & 0.95 & 0.99 & 1    &      &   &  \\ \hline
4192 & 20 & off & Recall@1    & 0.39 & 0.52 & 0.78 & 0.88 & 42  & 8h & 72.4M \\ \cline{4-8}
\rowcolor{LightGray}
      &    &     & Recall@1\%  & 0.91 & 0.95 & 0.99 & 1    &      &  &   \\ \hline
512  & 10 & off & Recall@1    & 0.39 & 0.51 & 0.97 & 0.90 &  95   & 16h & 10.3M \\ \cline{4-8}
\rowcolor{LightGray}
      &    &     & Recall@1\%  & 0.92 & 0.95 & 1    & 1    &      &   &  \\ \hline
512  & 5  & off & Recall@1    & 0.31 & 0.45 & 0.89 & 0.82 &   230  & 35h & 10.3M \\ \cline{4-8}
\rowcolor{LightGray}
      &    &     & Recall@1\%  & 0.82 & 0.85 & 0.95 & 0.91 &      &   &  \\ \hline
512  & 1  & off & Recall@1    & 0.25 & 0.40 & 0.71 & 0.69 &   1800  & 52h & 10.3M \\ \cline{4-8}
\rowcolor{LightGray}
      &    &     & Recall@1\%  & 0.70 & 0.79 & 0.91 & 0.99 &      &   &  \\ \hline
\textbf{512}  & \textbf{20} & \textbf{on}  & Recall@1    & \textbf{0.39} & \textbf{0.56} & \textbf{0.84} & \textbf{0.91} & 40 & 6h & 10.3M \\ \cline{4-8}
\rowcolor{LightGray}
      &    &     & Recall@1\%  & \textbf{0.96} & \textbf{0.96} & \textbf{1}    & \textbf{1}    &      &  &   \\ \hline
\end{tabular}
}
\end{table*}

\end{document}